\documentclass[a4paper,10.8pt]{book}

\usepackage[utf8]{inputenc}
\usepackage{geometry}
\usepackage[hidelinks]{hyperref}
\usepackage{titlesec}
\usepackage{etoolbox}
\usepackage{chngcntr}
\usepackage{graphicx}
\usepackage{lipsum}
\usepackage{xcolor}
\usepackage{amsmath,amssymb,amsfonts}
\usepackage{fancyhdr}
\usepackage{booktabs}
\usepackage{algorithm} 
\usepackage{algpseudocode}
\usepackage{setspace}
\usepackage{wrapfig}
\usepackage[parfill]{parskip}

\counterwithout{figure}{chapter}

\patchcmd{\thebibliography}{\chapter*}{\section*}{}{}

\titleformat{\chapter}[frame]{\normalfont}{}{10pt}{\LARGE\bfseries\filcenter}
\hypersetup{
    colorlinks=true,
    linkcolor=blue,
    filecolor=magenta,      
    urlcolor=cyan,
    }

\begin{document}
\begin{titlepage}
\chapter{Self-Driving Car Racing: Application of Deep Reinforcement Learning}

\thispagestyle{empty}
\vspace{-0.5cm}
\begin{center}
\begin{large}
		Florentiana Yuwono,
		Gan Pang Yen
		and Jason Christopher\\
		\vspace{0.25cm}
		\normalsize \; A0244109L, A0253516H, A0244120Y \\
        \normalsize \; \{e0851439, e0959104, e0851450\}@u.nus.edu \\
        \normalsize CS3263 Group 9\\ 
        \vspace{0.1cm}
\end{large}
\end{center}
\vspace{-0.5cm}
\end{titlepage}
\clearpage
\pagenumbering{arabic}
\section{Problem Understanding and Formulation}

\subsection{Motivation and rationale}
The motivation behind this project stems from the growing interest in autonomous driving systems, as evidenced by recent advancements in "smart car" technology, notably exemplified by companies such as Tesla. Furthermore, the landscape of AI-driven mobility has evolved towards implementing efficient systems for autonomous car racing, evident through new events such as Abu Dhabi Autonomous Racing League \cite{Baldwin2023} and the Indy Autonomous Challenge \cite{BusinessWire2024}, both happening in 2024. These groundbreaking events underscores the urgency for robust algorithms that can adeptly navigate dynamic environments.

By leveraging RL techniques, we seek to develop an AI agent capable of learning to drive a car efficiently in a simulated environment, given partial knowledge of the environment it is in. This research has implications not only in the field of autonomous vehicles but also in areas such as robotics and control systems.

\subsection{Innovativeness}
Application of RL in the domain of car racing is relatively less explored compared to other various control tasks. We aim to demonstrate their demonstrate their effectiveness in a challenging and dynamic scenario. We prioritize responsible AI considerations, which includes designing reward functions that prioritize safety as well as exploring techniques for interpretability and transparency in the learned policy.
We explore advanced RL algorithms such as DQN, PPO, and Transfer Learning integration to determine which yields the best results.

\subsection{Problem definition}

The problem we address is training an AI agent to effectively control a car in the \href{https://gymnasium.farama.org/environments/box2d/car_racing/}{OpenAI Gymnasium CarRacing environment}. This involves learning a policy that maps observations of the environment (camera images representing the car’s view of the track) to actions (steering, acceleration, and braking) in order to navigate the track while maximizing a performance metric (completing laps quickly without crashing). The objective is to achieve high performance in terms of both speed and safety, i.e. following the track, demonstrating the ability of RL algorithms to learn complex behaviors in dynamic environments.

\subsubsection{Observation space and stating state}

\begin{wrapfigure}{r}{0.4\textwidth} 
    \centering
    \includegraphics[width=0.4\textwidth]{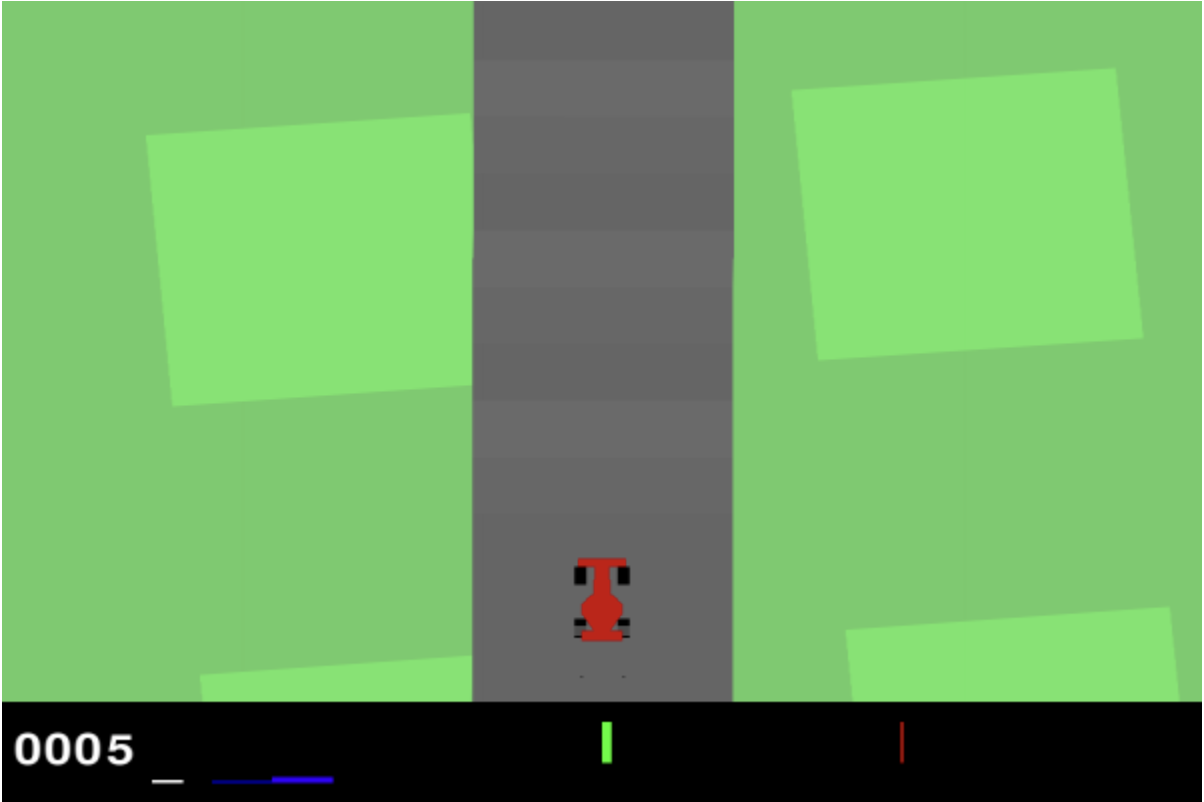}
    \caption{Game environment}
\end{wrapfigure}

The game environment consists frames of game state, where each frame is a $96\times 96$ RGB image of the car and race track, represented as \texttt{Box(0, 255, (96, 96, 3), uint8)} in Gymnasium package.

\subsubsection{Action space}
In discrete there are 5 actions: 0 = do nothing, 1 = full steer left, 2 = full steer right, 3 = full gas, 4 = full brake, represented as an int as indicated. 

In continuous there are 3 actions: steering (-1 is full left, +1 is full right), gas and breaking, represented as \texttt{Box([-1, 0, 0], 1.0, (3,), float32)}, a 3-dimensional array where action[0] = steering direction, action[1] = \% gas pedal and action[2] = \% brake pedal.

\subsubsection{Rewards and implied goal}

The agent will receive reward -0.1 every frame and +1000/$N$ for every track tile visited, where $N$ is the total number of tiles visited in the track. The goal is to finish the race successfully in the least number of frames possible (fast). We define “solving” as having an average reward of 800 over 100 consecutive trials.

\subsubsection{Starting state and episode termination}

The car starts at rest in the center of the road. Episode finishes when all the tiles are visited, or go off track and die with -100 reward.

\section{Knowledge and Technical Depth}

\subsection{Why reinforcement learning?}

Reinforcement Learning (RL) is particularly effective in scenarios such as CarRacing, where an agent's ability to learn through direct interaction with the environment and iterative feedback is crucial. RL also excels in deriving complex policies from direct sensory inputs, such as pixels, eliminating the need for predefined features. Conversely, alternative non-RL strategies, such as physics-based modeling and optimization techniques, provide enhanced computational efficiency and improved interpretability by leveraging established dynamics for analytical or simulation-based policy development. However, these methods depend heavily on domain-specific knowledge and may lack robustness in unfamiliar settings. 

\subsection{First step: formulate the problem as MDP}

To solve it with RL, the problem is first formulated as a Markov decision process (MDP) problem, where outcomes are partly random and partly under the control of the agent. The goal is to discover an optimal policy, denoted as $\pi^*$, which is a strategy for the agent that maximizes the expected cumulative reward over time. In this project, to find $\pi^*$, both value-based and policy-based methods are adapted. 

\subsubsection{Value-based methods}

This method tries to approximate the optimal action-value function (Q-function) given by:
\[
Q^{*} (s, a) = \max\limits_{\pi} \mathbb{E} [R_t|s_t = s, a_t = a, \pi]
\]
which assesses the expected return of taking a certain action in a given state. The agent then selects the action that has the highest expected return according to the Q-function.
The models implemented here include Deep Q-network and its self-customized variants, i.e. ResNet transfer learning and LSTM-ResNet variant.

\subsubsection{Policy-based methods}
This method directly parameterizes and learns the policy that maps states to actions without explicitly learning a value function. The model implemented here is Proximal Policy Optimization. 

\subsection{Deep Q-Network (DQN)}
DQN approximates the  Q-function with a deep neural network, i.e. $Q(s, a;\theta) \approx Q^*(s, a)$. Similar to the functionality of Q-table in Q-learning, Q-network takes in a state as input and outputs the predicted Q-values for each action, given a state. Then, it will store the agent's experience at each time-step in replay buffer and randomly samples a subset of the experience for training. 

\begin{figure}[ht]
    \centering
    \includegraphics[scale=1.3]{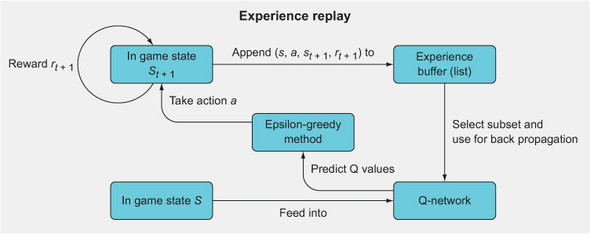}
    \caption{DQN. Source: https://livebook.manning.com/concept/deep-learning/q-network}
\end{figure}

\subsubsection{Full algorithm}

The core of the DQN algorithm is encapsulated in its loss function for the training of the Q-network. The loss function, $L_i(\theta_i)$, quantifies the difference between the predicted Q-value and the target Q-value. It's given by the mean squared error\cite{Mnih2013}:

\[
L_i (\theta_i) = \mathbb{E}_{s,a \backsim \rho(\cdot)} [ (y_i - Q (s, a; \theta_i))^2], \space \text{with target Q-value } y_i = \mathbb{E}_{s' \backsim \varepsilon} [r + \gamma \max_{a'} Q(s', a';\theta_{i-1})|s, a]
\]

To optimize the Q-function, stochastic gradient descent, namely RMSprop, is applied to the loss function (equation 3 in algorithm below): 

\[
\nabla_{\theta_i}L_i(\theta_i)=
\mathbb{E}_{s,a \backsim \rho(\cdot);s'\backsim\epsilon} 
[(r + \gamma \max_{a'} 
Q(s',a';\theta_{i-1}) - Q(s,a;\theta_i)) 
\nabla_{\theta_i}Q(s,a;\theta_i)]
\]

\begin{figure}[ht]
    \centering
    \includegraphics[scale=0.40]{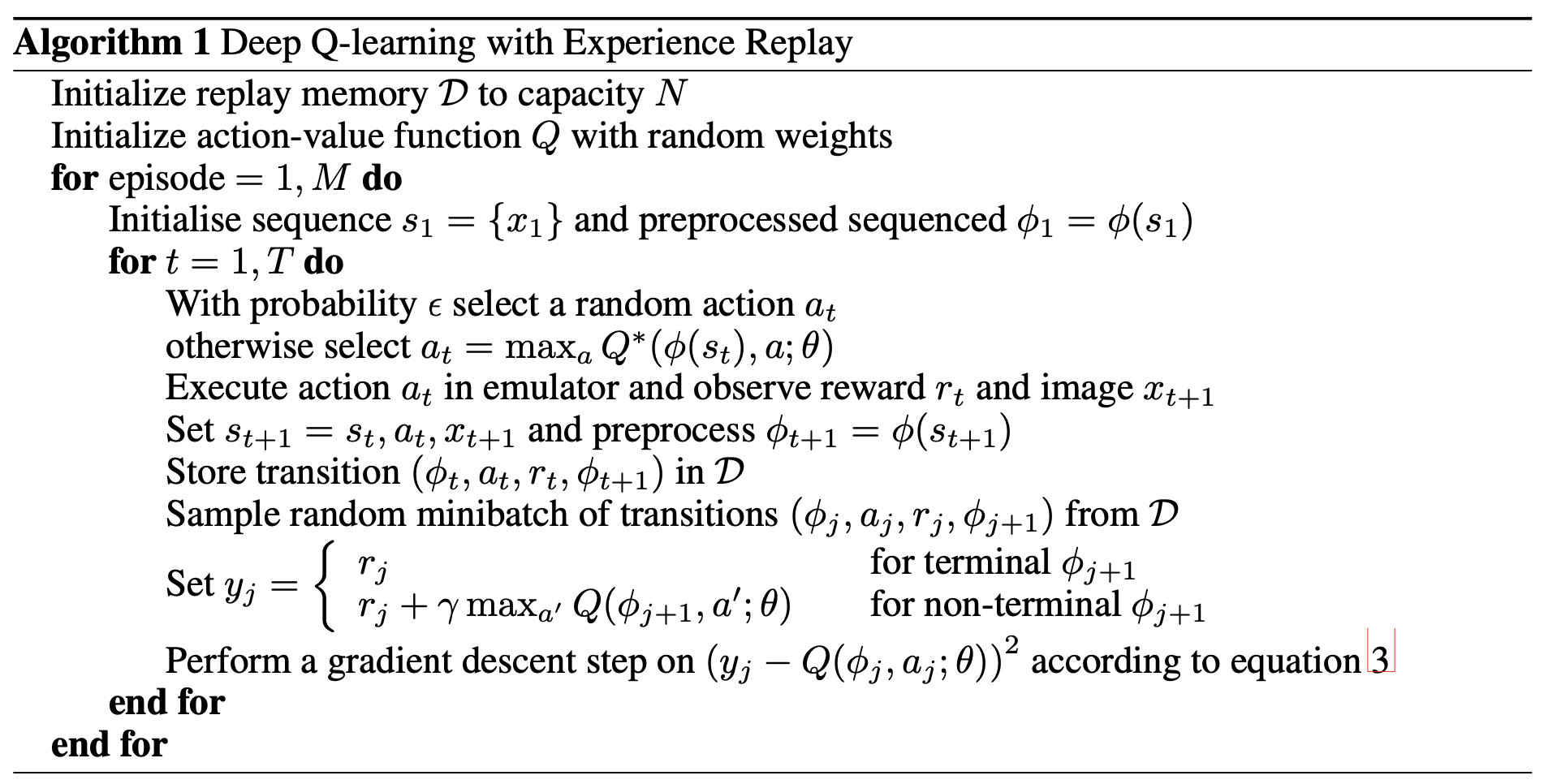}
    
    \caption{DQN algorithm. Source: https://arxiv.org/abs/1312.5602v1}
\end{figure}

\subsubsection{Advantages of DQN}
DQN can handle large state space with raw sensory inputs, such as images or complex state representations. The target network provides a stable target for the online network to learn from, while experience replay reduces the correlation between consecutive samples and helps to break the temporal dependencies and stabilize the learning\cite{Mnih2015}. 

\subsubsection{Improving DQN by Introducing Variants}
The original DQN tends to overestimate Q-values due to its maximization step in the Bellman equation update. To address this, the Double Q-learning variant\cite{Hasselt2016} was introduced to reduce the positive bias by separating action selection and evaluation in the Q-value update. Meanwhile, Prioritized Experience Replay\cite{Schaul2015} improves sample efficiency by prioritizing "surprising" experiences based on temporal difference error. Lastly, Deep Exploration via Bootstrapped DQN\cite{Osband2016} promotes a more effective exploration using an ensemble of Q-networks to gauge uncertainty in action selection.

\subsection{Transfer Learning and Recurrent Neural Networks}

Transfer learning is the reuse of a pre-trained model on a new problem. Transfer learning aims at improving the performance of target learners on target domains by transferring the knowledge contained in different but related source domains \cite{Zhuang2020}. While they have been extensively studied for supervised learning, it is an emerging topic for reinforcement learning \cite{Zhu2023}. There is a multitude of use cases for transfer learning, as it can help to process computer vision and natural language processing related tasks.

Recurrent Neural Networks (RNNs) are specific neural network architectures that detects patterns in sequential data \cite{Schmidt2019}. RNNs excel in its ability to capture temporal relationships in the data. Research have been done to integrate RNNs with DQN, which performs better due to the agent's ability to focus on particular previous states that are deemed important for predicting the action in the current state \cite{Chen2016}.

\subsection{Proximal Policy Optimization (PPO)}
Policy gradient methods, the foundation of Trust Region Policy Optimization (TRPO) and PPO, compute an approximation of the policy gradient and integrate it into a stochastic gradient ascent approach \cite{Schulman2017}. The prevalent gradient estimator is typically formulated as:

\[
\hat{g} = \mathbb{\hat{E}}_{t} \left[ \nabla_{\theta} \log \pi_{\theta}(a_{t} | s_{t}) \hat{A}_{t} \right]
\]

Here, \(\pi_{\theta}\) represents a stochastic policy, and \(\hat{A}_{t}\) is an estimate of the advantage function at time step \(t\). The expectation \(\mathbb{\hat{E}}_{t}[\cdot]\) denotes the empirical average over a finite batch of samples, within an algorithm that iterates between sampling and optimization. Implementations employing automatic differentiation software create an objective function whose gradient yields the policy gradient estimator. The estimator \(\hat{g}\) is derived by differentiating the objective:

\[
\mathcal{L}^{PG}(\theta) = \mathbb{\hat{E}}_{t} \left[ \log \pi_{\theta}(a_{t} | s_{t}) \hat{A}_{t} \right]
\]

While it may seem enticing to perform multiple optimization steps on this loss \(\mathcal{L}_{PG}\) using the same trajectory, such an approach lacks sufficient justification. Empirically, it often results in excessively large policy updates.

TRPO maximizes a surrogate objective while adhering to a constraint on the magnitude of the policy update. This optimization problem is formulated as:

\[
\max_\theta \mathbb{\hat{E}}_{t} \left[ \frac{\pi_{\theta}(a_{t} | s_{t})}{\pi_{\theta_{\text{old}}}(a_{t} | s_{t})} \hat{A}_{t} \right] \text{subject to } \mathbb{\hat{E}}_{t} \left[ \text{KL}[\pi_{\theta_{\text{old}}}( \cdot | s_{t}), \pi_{\theta}( \cdot | s_{t})] \right] \leq \delta
\]

Here, \(\theta_{\text{old}}\) represents the vector of policy parameters before the update. TRPO utilizes a penalty instead of a strict constraint in an unconstrained optimization problem to maintain monotonic improvement. However, selecting a suitable value for the penalty coefficient (\(\beta\)) poses challenges in generalization across different problems.

PPO addresses the limitations of TRPO by introducing a clipped surrogate objective:

\[
\mathcal{L}^{\text{CLIP}}(\theta) = \mathbb{\hat{E}}_{t} \left[ \min \left( r_t(\theta) \hat{A}_t, \text{clip} \left( r_t(\theta), 1 - \epsilon, 1 + \epsilon \right) \hat{A}_t \right) \right]
\]

where \(\epsilon\) is a hyperparameter (e.g., \(\epsilon = 0.2\)). This objective ensures that policy updates remain within a reasonable range by constraining the probability ratio. By choosing the minimum between the clipped and unclipped objectives, PPO maintains a lower bound on the unclipped objective, thus penalizing excessively large updates.

\begin{algorithm}[ht]
\caption{PPO, Actor-Critic Style \cite{Schulman2017}}
\begin{algorithmic}
    \For{$\text{iteration} = 1, 2, \ldots$}
        \For{$\text{actor} = 1, 2, \ldots, N$}
            \State Run policy $\pi_{\theta_{\text{old}}}$ in environment for $T$ timesteps
            \State Compute advantage estimates $\hat{A}_1, \ldots, \hat{A}_T$
        \EndFor
        \State Optimize surrogate $L$ with respect to $\theta$, with $K$ epochs and minibatch size $M \leq NT$
        \State $\theta_{\text{old}} \gets \theta$
    \EndFor
\end{algorithmic}
\end{algorithm}

Advantages: PPO offers simplicity in implementation, greater generality, and improved empirical stability and data efficiency. Notably, it performs well on continuous action spaces (where DQN struggles \cite{WangK}) and doesn't require extensive hyperparameter tuning.

\section{Methodology and Results}

\subsection{Preprocessing}

\subsubsection{Modify \texttt{env.reset()} to always skip the first 50 states}

The game configuration always just gradually zooms in for the first 50 steps. Since this zoom-in phase is a very small part of the overall game, keeping this during training might hinder our agent from learning to control the car in the main frames after zoom-in.

\subsubsection{Convert the image to grayscale and resize to $84\times 84$}

This is to reduce the number of dimension (3 channels to 1) and allow for more compact states. For example truncating the black bar at the bottom of the frame and the both horizontal ends of the frame.

\subsubsection{Modify \texttt{env.step()} to use frame skipping technique}

The agent sees and selects actions on every 4th frame instead of every frame, and its last action is repeated on skipped frames. Hence, we can transform each observation to contain 4 frames simultaneously so that the agent can know whether it is moving forward or backward. This will also help to decrease the time since we only need 1 action per 4 frames.

\begin{figure}[ht]
    \centering
    \includegraphics[scale=0.4]{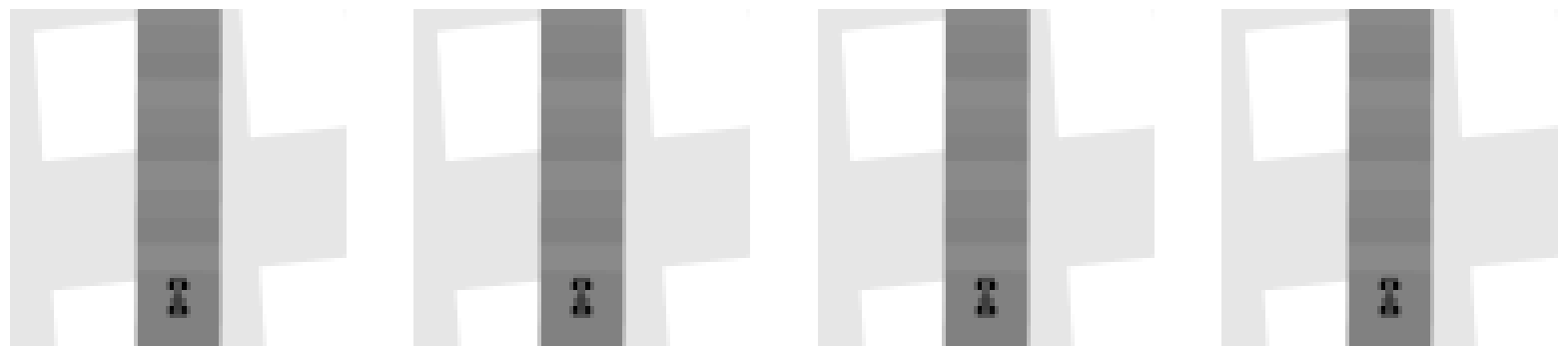}
    \caption{Shape of the observation after preprocessing: (4, 84, 84)}
\end{figure}

\subsection{Implementation and result of DQN \href{https://github.com/pangyyen/carRacing-DeepRL/blob/main/dqn/main_dqn_training.ipynb}{[GitHub link]}}
We implemented DQN from scratch by defining the Deep Neural Network Class used in DQN, Experience Replay buffer Class, DQN agent, and implemented the training algorithm which saves the model and evaluates it every 10,000 time steps. In addition, we have also implemented an epsilon decay rate to maximize exploration at the beginning of the training and gradually shift towards exploitation over the course of training. 

Eventually, the algorithm took 15 hours to reach the max average of performance, 910.12 after 1.45 million time steps, and started to oscillate afterwards. The full training process is shown in Figure \ref{fig:dqn_training}:
\begin{figure}[ht]
    \centering
    \includegraphics[scale=0.3]{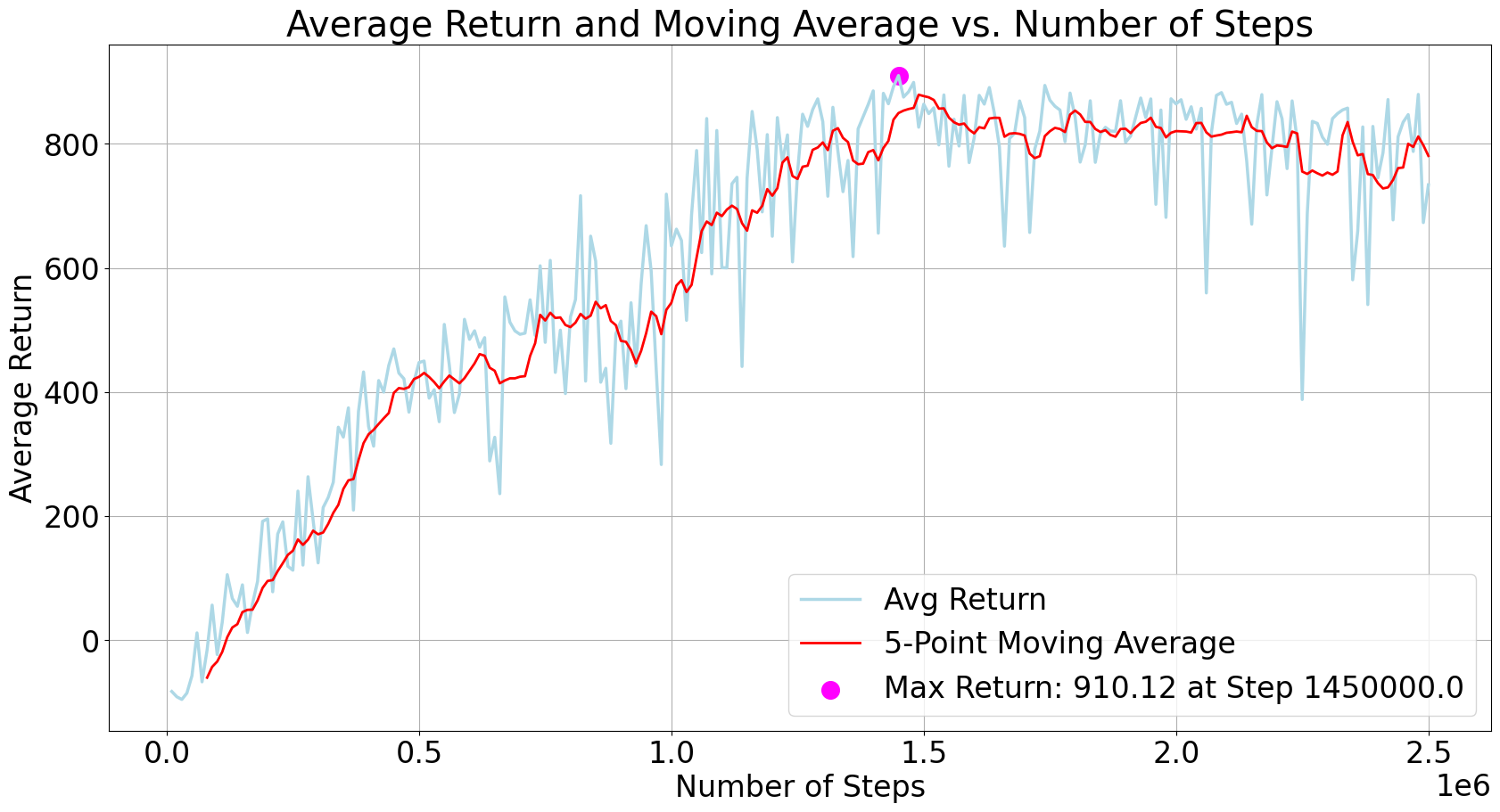}
    \caption{DQN training process: \href{https://github.com/pangyyen/carRacing-DeepRL/blob/main/dqn/main_dqn_analysis.ipynb}{[GitHub link]}}
    \label{fig:dqn_training}
\end{figure}

\subsection{Implementation of Transfer Learning \href{https://github.com/pangyyen/carRacing-DeepRL/blob/main/resnet/ResNet.ipynb}{[GitHub link]}}

We improved upon the classical DQN implementation by replacing the first two convolutional layers with PyTorch's ResNet-18 pretrained model, which incorporates deep residual learning for image recognition. We chose ResNet-18 as it is relatively lightweight compared to other variants such as ResNet-50 or ResNet-101, making it suitable for reinforcement learning training.

We changed the preprocessing stage to take in all three RGB color channels to fit the ResNet-18 input. Our model will now process one image frame at a time.

\begin{figure}[ht]
    \centering
    \includegraphics[scale=0.35]{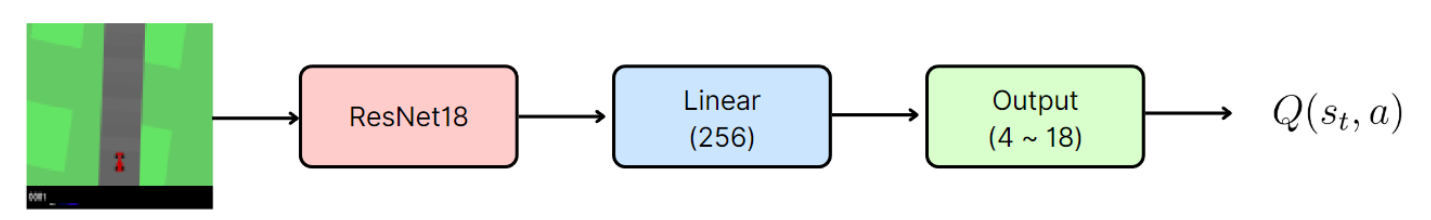}
    \includegraphics[scale=0.35]{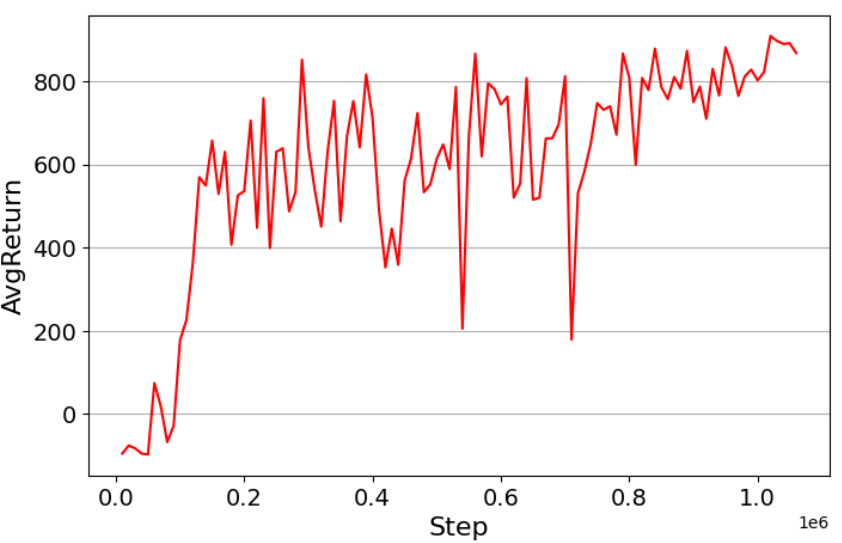}
    \caption{Training performance with transfer learning}
\end{figure}

Our model was trained on Google Colab's L4 GPU for 23 hours. We observed that as compared to the DQN + CNN implementation, the model seemed to learn in relatively fewer steps, reaching an average return of 600 in less than 200,000 time steps. This model reached a peak performance of 912 after 1,200,000 time steps. This performance is likely contributed by the image segmentation effect produced by the ResNet layer, capturing more meaningful spatial relationships as compared to traditional CNN implementation.

\subsection{Implementation of Transfer Learning + RNN Combination \href{https://github.com/pangyyen/carRacing-DeepRL/blob/main/resnet-lstm/ResNet-LSTM.ipynb}{[GitHub link]}}

To overcome the problem of capturing temporal relationships, we propose a new method of combining transfer learning along with sequential models for reinforcement learning.

\begin{figure}[ht]
    \centering
    \includegraphics[scale=0.4]{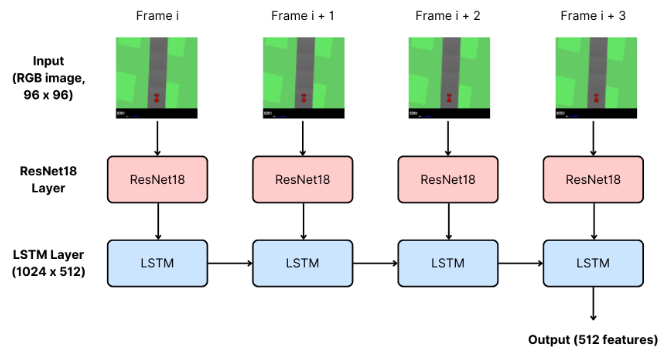}
    \includegraphics[scale=0.4]{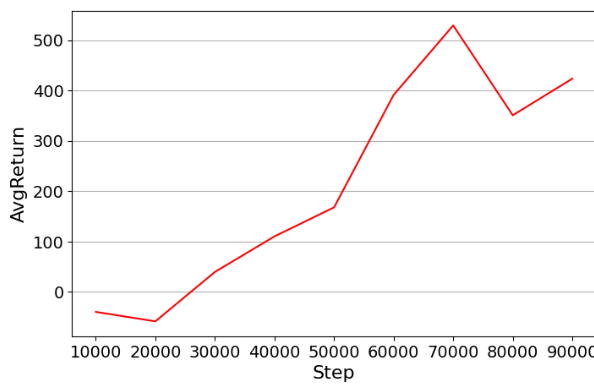}
    \caption{ResNet-LSTM training performance}
    \label{fig:resnet-lstm}
\end{figure}

This method replaces the ResNet-18 image recognition layer with a combined ResNet-LSTM layer, which connects each ResNet-18 layer with a LSTM cell as shown in Figure \ref{fig:resnet-lstm}.

We observe that the contribution of spatio-temporal relationship through the combination of an image segmentation and a memory layer contributed to a faster convergence to reach high average return values in less than 100,000 steps. 

However, this approach demands substantially greater computational resources in contrast to alternative methods. This limitation prompted our team to discontinue model evaluation after 90,000 time steps. Our model was trained on Google Colab's A100 GPU for 8 hours and surpassed Colab's 83.5 GB system RAM at this iteration count. Possible improvements may include improving computational efficiency by reducing model parameter size, or by applying distributed algorithms to reinforcement learning \cite{Kapturowski2019}

\subsection{Implementation of PPO}

\subsubsection{PPO on discrete action space \href{https://github.com/pangyyen/carRacing-DeepRL/blob/main/ppo/ppo_discrete.ipynb}{[GitHub link]}}

\begin{figure}[ht]
    \centering
    \includegraphics[scale=0.36]{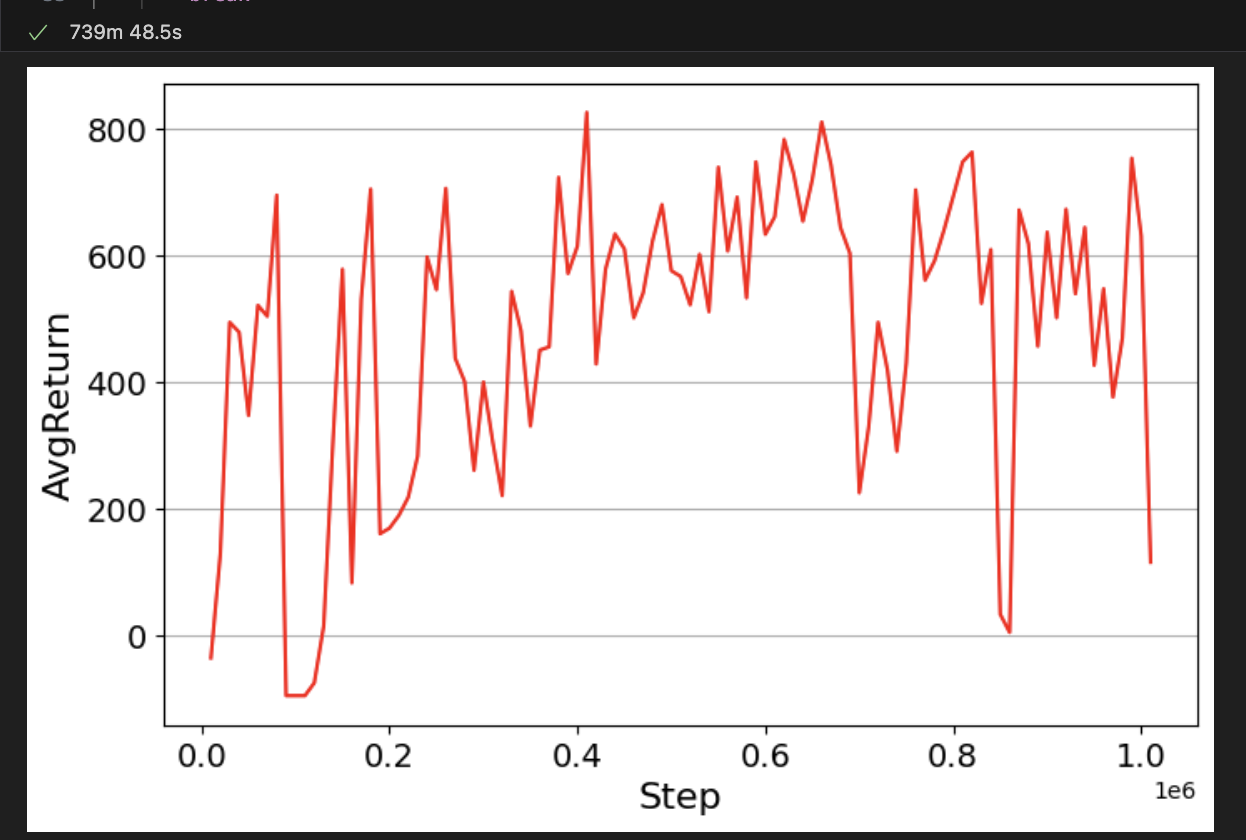}
    \includegraphics[scale=0.35]{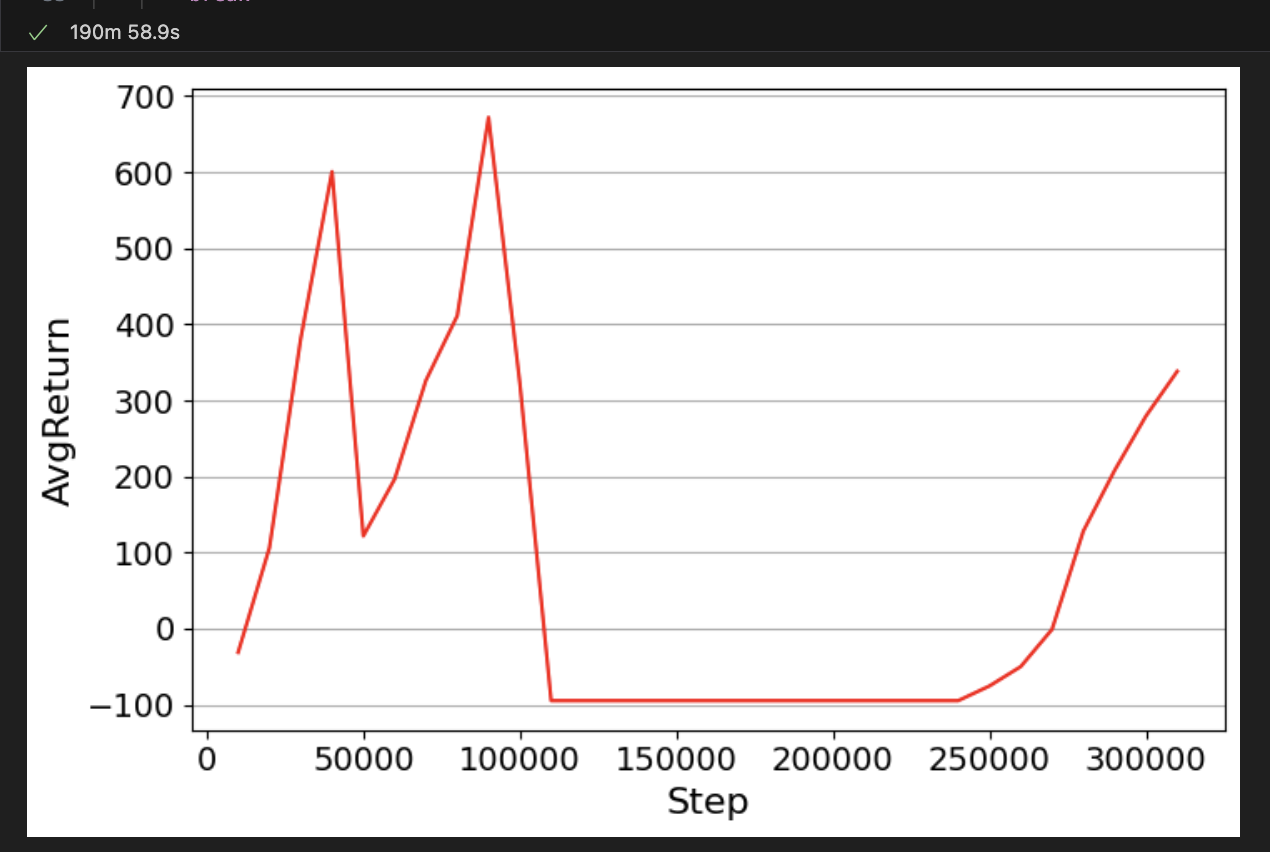}
    \caption{PPO with default Adam (left) vs. non-stationary Adam (right)}
    \label{fig:ppo-discrete}
\end{figure}

We customized stable-baselines3 implementation of PPO for training the environment and observed that the learning happened faster than DQN on the first 400K timesteps, reaching an average score of 800. This performance became more stable with smaller deviation until 600K timesteps. However, after that the performance suddenly became unstable, as indicated in Figure \ref{fig:ppo-discrete},

We suspected that this is due to a phenomenon called policy collapse, where as agent continues to interact with the environment, their performance degrades. Research \footnote{Shibhansh Dohare, Qingfeng Lan, A. Rupam Mahmood, "Overcoming Policy Collapse in Deep Reinforcement Learning", Published: 20 Jul 2023, Last Modified: 29 Aug 2023.} has shown that the standard use of Adam can lead to sudden large weight changes even when the gradient is small whenever there is non-stationarity in the data stream.

We followed the paper to customize Adam optimizer with equal values of betas at 0.99, and achieved the performance where the model can learn the policy very quickly (beat others in reaching 700 in less than 100K steps), but also collapse very quickly. We suspected that the values of betas are still subject to hyperparameter tuning. We think that this phenomenon of policy collapse serves as an interesting area to explore in future research work.

\subsubsection{PPO on continuous action space \href{https://github.com/pangyyen/carRacing-DeepRL/blob/main/ppo/ppo.ipynb}{[GitHub link]}}

We normalized the action space to be continuous between interval [-1, 1] due to the Gaussian distribution (mean=0, std=1) implementation for continuous actions. During the initial 350K time steps of training, we observed that the performance was still unstable, although there was upward trend in average return. This mainly can be caused due to huge continuous action space.

\begin{figure}[ht]
    \centering
    \includegraphics[scale=0.35]{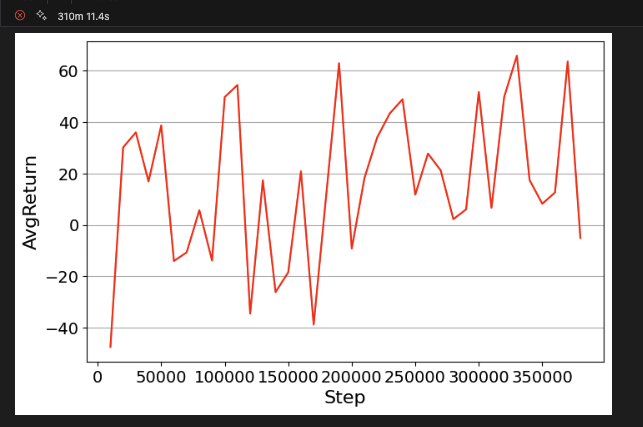}
    \caption{PPO training with continuous action space}
\end{figure}

We predict that PPO will still be able to achieve good performance as we increase the timesteps (as proven by several other research results \footnote{https://github.com/elsheikh21/car-racing-ppo}), however we are constrained by the compute power to prove this. Nevertheless, we still think that demonstrating this capability is important in real-world scenario of self-driving, where you should be able to do 20\% gas and 10 degree left turn (achievable by continuous action space), instead of only full gas or full left turn at a time (discrete action).

\subsection{Performance comparison}

\begin{figure}[ht]
    \centering
    \includegraphics[scale=0.035]{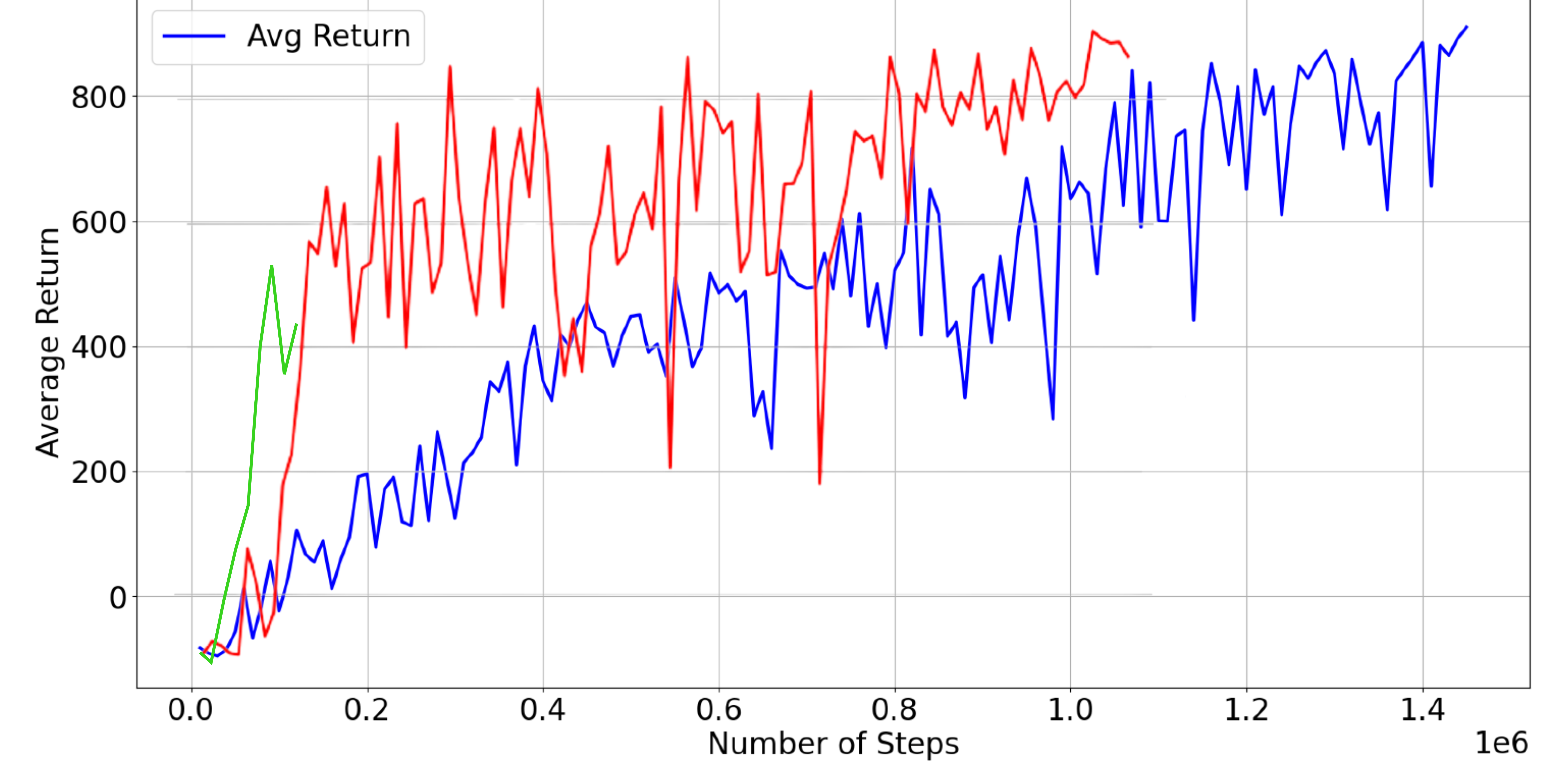}
    \includegraphics[scale=0.29]{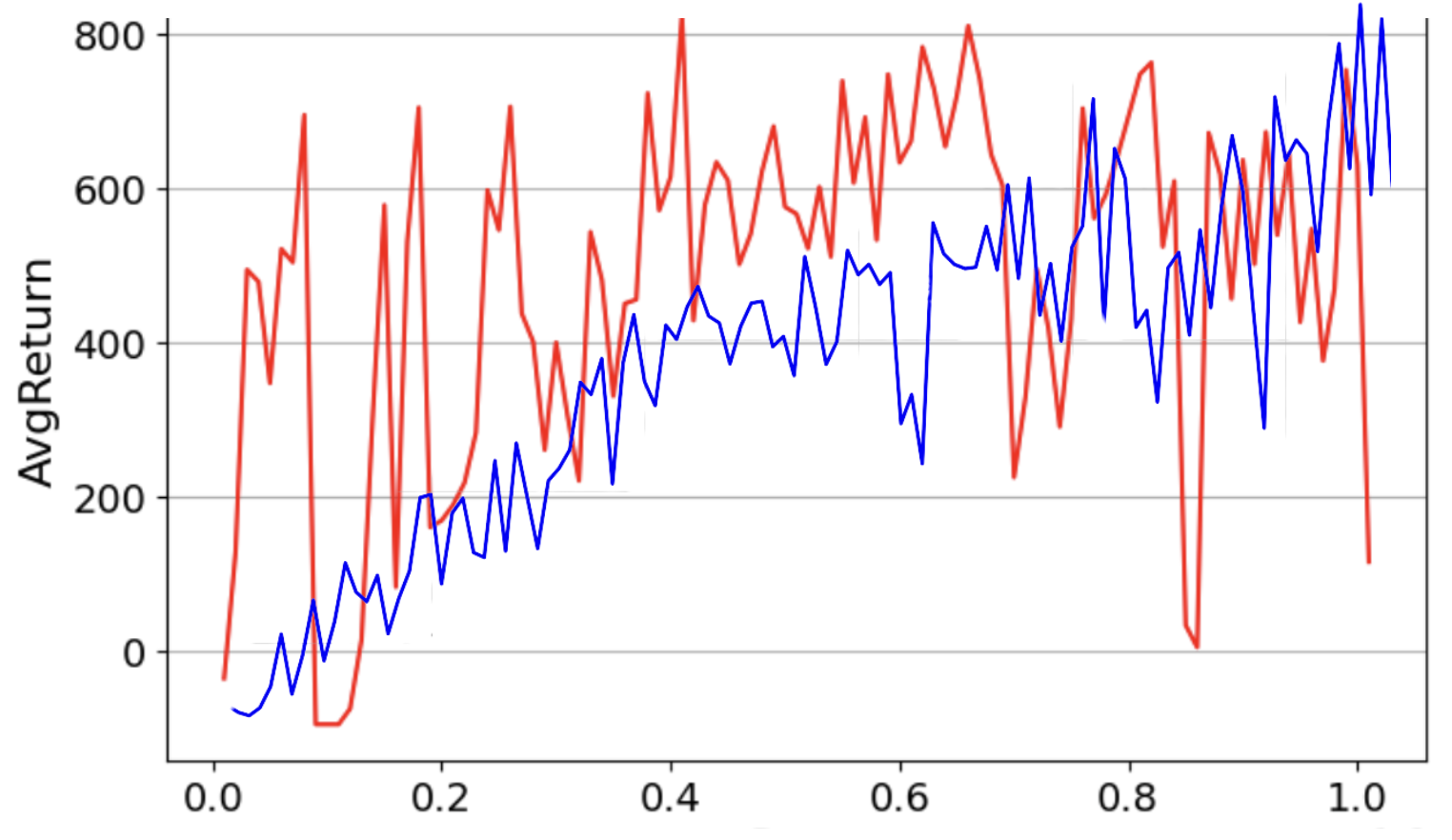}
    \caption{DQN(blue) vs ResNet(red) vs ResNet-LSTM(Green); and DQN(blue) vs PPO(red)}
    \label{fig:performance comparison}
\end{figure}

Notably, it can be seen from Figure \ref{fig:performance comparison} that incorporating transfer learning (ResNet) into DQN enhances the agent's performance since it allows the agent to grasp a more robust and informed representation of the environment. This allows the agent to achieve near-peak performance in far less iterations. We believe this result is driven by the presence of quality information, as replacing the convolutional layers by a pretrained ResNet layer allows the agent to better identify important features in the frame, allowing a more efficient learning iteration.

ResNet with LSTM also seems to provide promising results from the first few iterations, as we believe that the introduction of RNNs into the model allows the capturing of spatio-temporal relationships within and between frames. However, as the model learning was ended prematurely due to limited compute power, more research may be required to arrive at a better conclusion on the ResNet-LSTM performance. Another notable observation lies on the time taken per iteration. While we see that the transfer learning options achieves high performance in fewer iterations, due to the added complexity of the ResNet layer, each iteration now takes a longer time, taking on average 45 minutes per 10,000 steps, far longer than DQN-CNN iterations.

Whereas PPO can be seen to beat DQN in reaching higher average return in shorter time, it is much more unstable and sensitive to experiencing policy collapse during training. This can be due to PPO's reliance on a fixed-size trust region. When the policy deviates too much from the previous policy, the trust region constraint can lead to overly conservative updates, where the policy becomes stuck in suboptimal solution.

Comparing the performances of our AI agents with human players (ourselves) \footnote{\url{https://drive.google.com/drive/folders/1ntYOZsL1ZZ1l8miHlr2z3E1mUHlHdVwD?usp=sharing}}, the average score by these 3 human players are around 800, whereas the AI could reach an average of 850-900 reward consistently. We welcome human testers to play the game \href{https://github.com/pangyyen/carRacing-DeepRL/blob/main/keyboard_play.py}{here}.

\subsection{Model behaviour}
\subsubsection{Intermediate behaviour}
The videos uploaded to a \href{https://drive.google.com/drive/folders/1fWSdTtYF16-KPDrfU865zeV_7kaUR2MK}{ Google Drive folder} illustrated the exploration behavior of the model during training. The first video shows that the agent indeed treated all actions equally at the start of the training, causing it to struggle with going forward even though it was on a straight route. The second video shows that the agent has acquired the ability to drive decently fast on the track, despite consistently performing some minor turns along the way.

\subsubsection{Final behaviour}
The three demonstration videos in the google drive showcase the agent's advanced driving capabilities. In particular, the agent has learned to perform delicate drifting and handle skidding when encountering U-turns, a challenging maneuver especially at high speed when the car is prone to skidding. Furthermore, the agent demonstrates the ability to slow down appropriately when navigating sharp turns. On straight routes, the agent consistently applies the "gas" to maintain optimal speed. These behaviors highlight the agent's adaptability and its capacity to make intelligent decisions based on the track's layout, ultimately resulting in a smooth and efficient driving performance.

\begin{figure}
    \centering
    \includegraphics[scale=0.1]{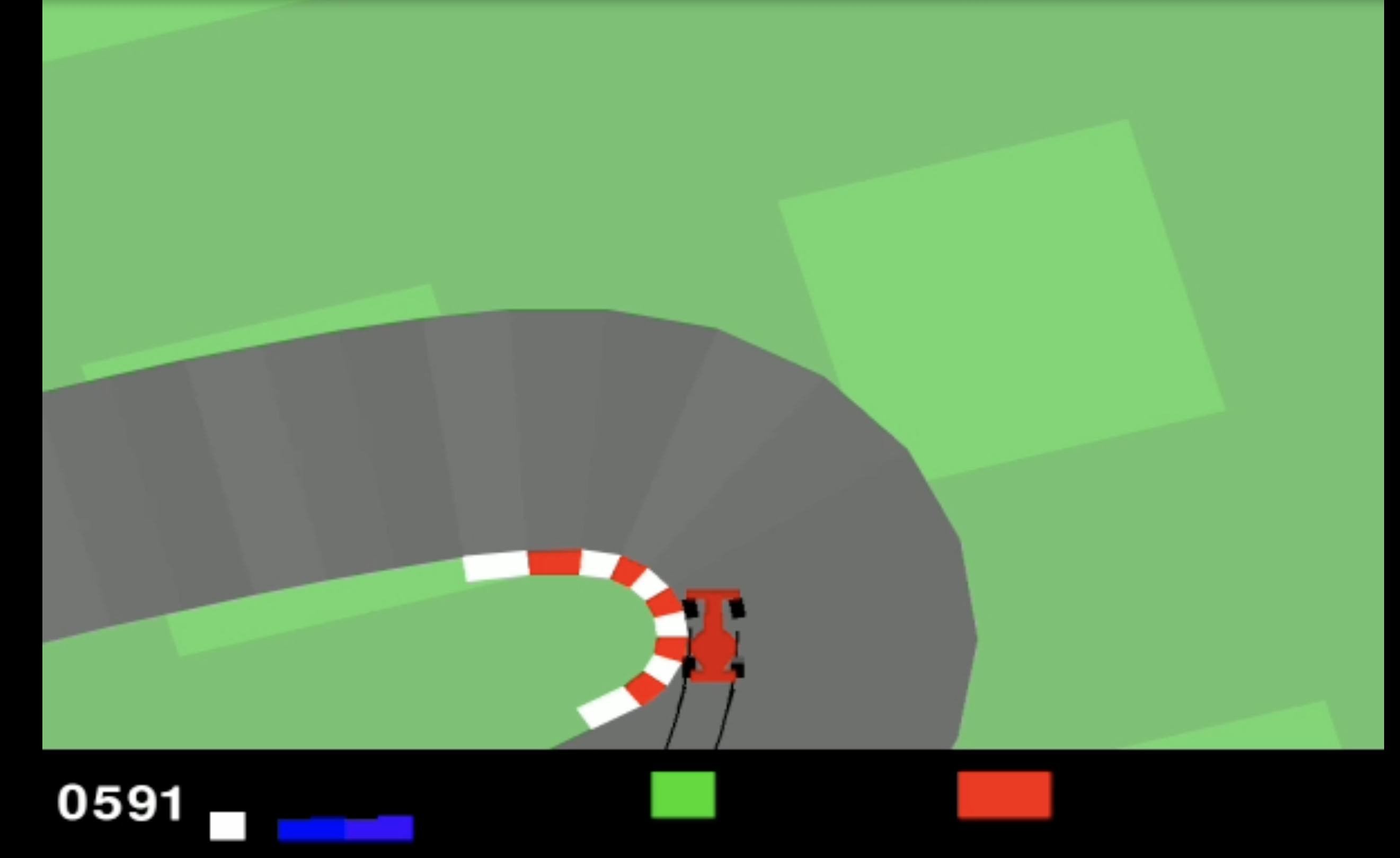}
    \includegraphics[scale=0.09]{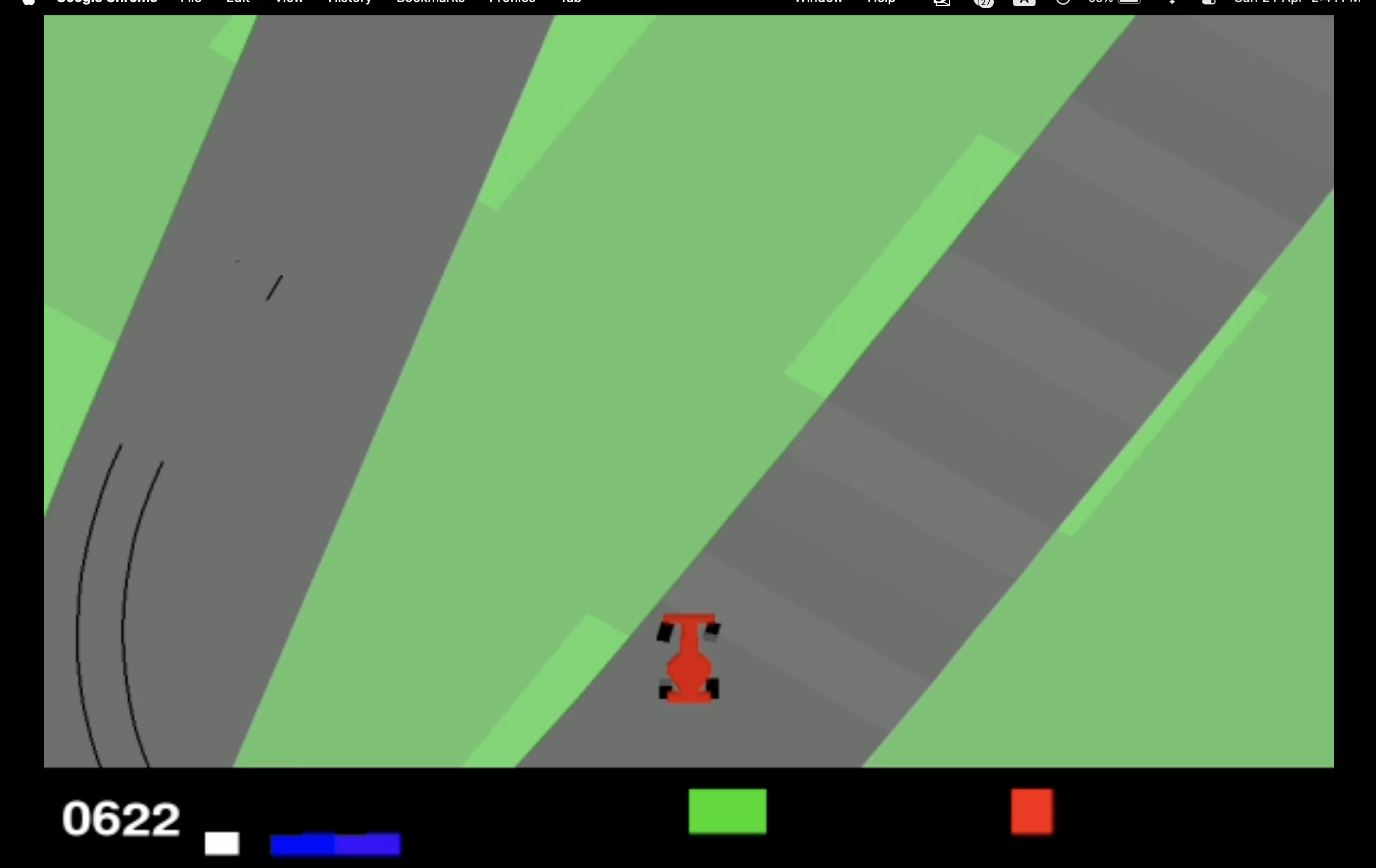}
    \caption{The agent being able to handle the potential skid well, i.e. drifting}
\end{figure}

\section{Effort and Initiative}

\subsection{Failed project: LuxAI Season 2}

We spent around 2-3 weeks trying to debug the LuxAI compatibility issues, due to its outdated implementation. But as we tried to fix one version, it becomes incompatible with other things (python versioning, stable-baselines, kaggle notebook, google colab, local machine).

\subsection{Evidence of effort and work}

We spent a lot of hours on reading research papers, other implementations, coded out the solutions, with more than 100 hours waiting for model training. We also had 2-3 meetings in a week.

\subsection{Team member contributions}

Each member contributes towards writing the proposal, report and presentation deck. Specific task allocations of each member are as follows:

{\bf Florentiana Yuwono}: overall direction of the team, research on papers and implementation, in charge of PPO implementation. \\
{\bf Gan Pang Yen}: research on papers and implementation, in charge of DQN and PPO training. \\
{\bf Jason Christopher}: research on papers and implementation, in charge of ResNet and ResNet + LSTM model design.

\section{Conclusion}


This project has demonstrated the potential and effectiveness of various deep reinforcement learning algorithms in navigating a car autonomously in a simulated environment. Through extensive experimentation with DQN, PPO, and innovative adaptations incorporating transfer learning and RNNs, we have uncovered significant insights into the strengths and limitations of each approach within the context of self-driving car racing.

Our findings reveal that while DQN provides a robust foundation, the incorporation of advanced neural network architectures like ResNet and LSTM can enhance the agent's performance by enabling it to capture complex spatial and temporal dependencies within the environment. Meanwhile, PPO has shown promising results, particularly in scenarios requiring fine control over continuous action spaces, which are crucial for realistic driving simulations. 

The integration of ResNet with LSTM, while offering superior ability to capture spatio-temporal relationships, poses significant computational challenges. To facilitate the scaling of such models to millions of time steps, further enhancements in computational efficiency or access to more substantial computing resources will be necessary. This could involve optimizing the architecture for better performance on available hardware or employing more advanced parallel computing techniques. Future work will focus on refining these models and exploring the integration of these techniques into actual autonomous driving systems. Additionally, further research into the phenomenon of policy collapse in PPO could lead to more stable and reliable learning algorithms.

This project not only advances our understanding of applying deep reinforcement learning to autonomous driving but also sets the stage for future innovations in this exciting and rapidly evolving field.

\pagebreak











\end{document}